\documentclass[runningheads]{llncs}
\titlerunning{SAAS}
\usepackage{graphicx}
\usepackage{amsmath}
\usepackage{comment}
\usepackage{algorithm}
\usepackage{algorithmic}
\usepackage{booktabs}
\usepackage{multirow}
\usepackage{subfigure}
\usepackage{xcolor}
\usepackage[misc]{ifsym}
\usepackage[colorlinks,
            citecolor=blue,
            linkcolor=blue,
            urlcolor=blue
            ]{hyperref}
\usepackage{epstopdf}
\usepackage{comment}
\begin{document}
%
%\title{SAAS: Selective Annotation and Aggregative Supervision for Active Learning on Skin Lesion Analysis}
%\title{Active Sample Selection and Aggregation for Annotation Reduction in Skin Lesion Analysis}

%\title{Active Sample Selection and Aggregation for Cost-effective Labelling in Skin Lesion Analysis！}
\title{An Active Learning Approach for Reducing Annotation Cost in Skin Lesion Analysis}

\titlerunning{An Active Learning Approach for Skin Lesion Analysis}
% If the paper title is too long for the running head, you can set
% an abbreviated paper title here
%

\author{Xueying Shi\textsuperscript{1(\Letter)}, Qi Dou\textsuperscript{2}, Cheng Xue\textsuperscript{1}, Jing Qin\textsuperscript{4}, Hao Chen\textsuperscript{1,3}, \\Pheng-Ann Heng\textsuperscript{1,5}}
\authorrunning{X. Shi et al.}
% First names are abbreviated in the running head.
% If there are more than two authors, 'et al.' is used.
%
\institute{\textsuperscript{\rm 1} Department of Computer Science and Engineering \\The Chinese University of Hong Kong, Hong Kong, China\\
\email{xyshi@cse.cuhk.edu.hk}\\
\textsuperscript{\rm 2} Department of Computing
Imperial College London, London SW7 2AZ, U.K.\\
\textsuperscript{\rm 3} Imsight Medical Technology Co., Ltd., Shenzhen, China\\
\textsuperscript{\rm 4} Centre for Smart Health, School of Nursing \\The Hong Kong Polytechnic University, Hong Kong, China\\
\textsuperscript{\rm 5}Guangdong Provincial Key Laboratory of Computer Vision and Virtual Reality Technology, Shenzhen Institutes of Advanced Technology, Chinese Academy of Sciences, Shenzhen, China}

\maketitle              % typeset the header of the contribution
\begin{abstract}
Automated skin lesion analysis is very crucial in clinical practice, as skin cancer is among the most common human malignancy. Existing approaches with deep learning have achieved remarkable performance on this challenging task, however, heavily relying on large-scale labelled datasets.
In this paper, we present a novel active learning framework for %annotation cost-effective 
cost-effective skin lesion analysis.
The goal is to effectively select and utilize much fewer labelled samples, while the network can still achieve state-of-the-art performance.
Our sample selection criteria complementarily consider both informativeness and representativeness, derived from decoupled aspects of measuring model certainty and covering sample diversity.
To make wise use of the selected samples, we further design a simple yet effective strategy to aggregate intra-class images in pixel space, as a new form of data augmentation.
We validate our proposed method on data of \emph{ISIC 2017 Skin Lesion Classification Challenge} for two tasks.
Using only up to 50\% of samples, our approach can achieve state-of-the-art performances on both tasks, which are comparable or exceeding the accuracies with full-data training, and outperform other well-known active learning methods by a large margin.
\end{abstract}

\section{Introduction}
\begin{comment}
Deep convolutional networks have revolutionized medical image recognition in recent years, achieving even expert-level performance on various challenging tasks, e,g., skin lesion analysis~\cite{esteva2017dermatologist}. However, these achievements heavily rely on accurate and extensive annotations of medical images, which is tedious, time-consuming and skill-demanding.
\end{comment}

Skin cancer is among the most common cancers worldwide, and accurate analysis of dermoscopy images is crucial for reducing melanoma deaths~\cite{esteva2017dermatologist}.
Existing deep convolutional neural networks (CNNs) have demonstrated appealing efficacy for skin lesion analysis, even setting dermatologist-level performance.
%such as benign/malignant classification and seborrheic keratosis/non seborrheic keratosis classification.
However, these achievements heavily rely on extensive labelled datasets, which is very expensive, time-consuming and skill-demanding.
Recently, with increasing awareness of the impediment from unavailability of large-scale labeled data, researchers have been frequently revisiting the concept of active learning to train CNNs in a more cost-effective fashion~\cite{settles.tr09}.
The goal is to learn CNNs with much fewer labelled images, while the model can still achieve the state-of-the-art performance.

Sample selection criteria usually use informativeness or representativeness~\cite{huang2010active}.
%On the one hand, samples receiving low-confidence predictions are regarded as informative for further model updating.
Informative samples are the ones which the current model still cannot recognize well.
For example, Mahapatra et al.~\cite{mahapatra2018efficient} derived uncertainty metrics via a Bayesian Neural Network to select informative samples for chest X-ray segmentation.
On the other hand, representativeness measures whether the set of selected samples are diverse enough to represent the underlying distributions of the entire data space. 
Zheng et al.~\cite{aaai2019ra} chose representative samples with unsupervised feature extraction and clusters in latent space.
Moreover, rather than only relying on one single criterion, some works actively select samples by integrating both criteria.
Yang et al.~\cite{yang2017suggestive} selected samples which receive low prediction probabilities and have large distances in CNN feature space.
Another state-of-the-art method is AIFT~\cite{zhou2017fine} (active, incremental fine-tuning), which employed the entropy of CNN predictions for a sample to compute its informativeness as well as representativeness, demonstrating effectiveness on three medical imaging tasks.
However, these existing methods derive both criteria based on the same CNN model, which hardly avoid potential correlations within the selected samples. How to exploit such dual-criteria in a more disentangled manner still remains open.

With active sample selection, the data redundancy of unlabelled sample pool is effectively reduced. Meanwhile, we should note that the obtained set of images come with high intra-class variance in color, texture, shape and size~\cite{xue2019robust,zhang2018skin}. Directly using such samples to fine-tune the model may fall into more-or-less hard example mining, and face the risk of over-fitting.
Hence, we argue that it is also very critical to more wisely use the compact set of selected samples, for unleashing their value to a large extent.
However, sample utilization strategies receive less attention in existing active learning literatures.
One notable method is mix-up~\cite{zhang2017mixup}, that augments new training data as pixel-wise weighted addition of two images from different classes. However, mix-up is not suitable for situations where data have large intra-class variance while limited inter-class variance, which is exactly our case at skin lesion analysis.

In this work, we propose a novel active learning method for skin lesion analysis to improve annotation efficiency. Our framework consists of two components, i.e., sample selection and sample aggregation.
Specifically, we design dual-criteria to select informative as well as representative samples, so that the selected samples are highly complementary.
Furthermore, for effective utilization of the selected samples, we design an aggregation strategy by augmenting intra-class images in pixel space, in order to capture richer and more distinguishable features from these valuable yet ambiguous selected samples.
We validate our approach on two tasks with the dataset of \emph{ISIC 2017 Skin Lesion Classification Challenge}.
We achieve state-of-the-art performance by using 50\% data for task 1 and 40\% for task 2 of skin lesion classification tasks.
In both tasks, our proposed method consistently outperforms existing state-of-the-art active learning methods by a large margin.

\begin{figure}[t]
    \centering
    \includegraphics[width=0.75\linewidth]{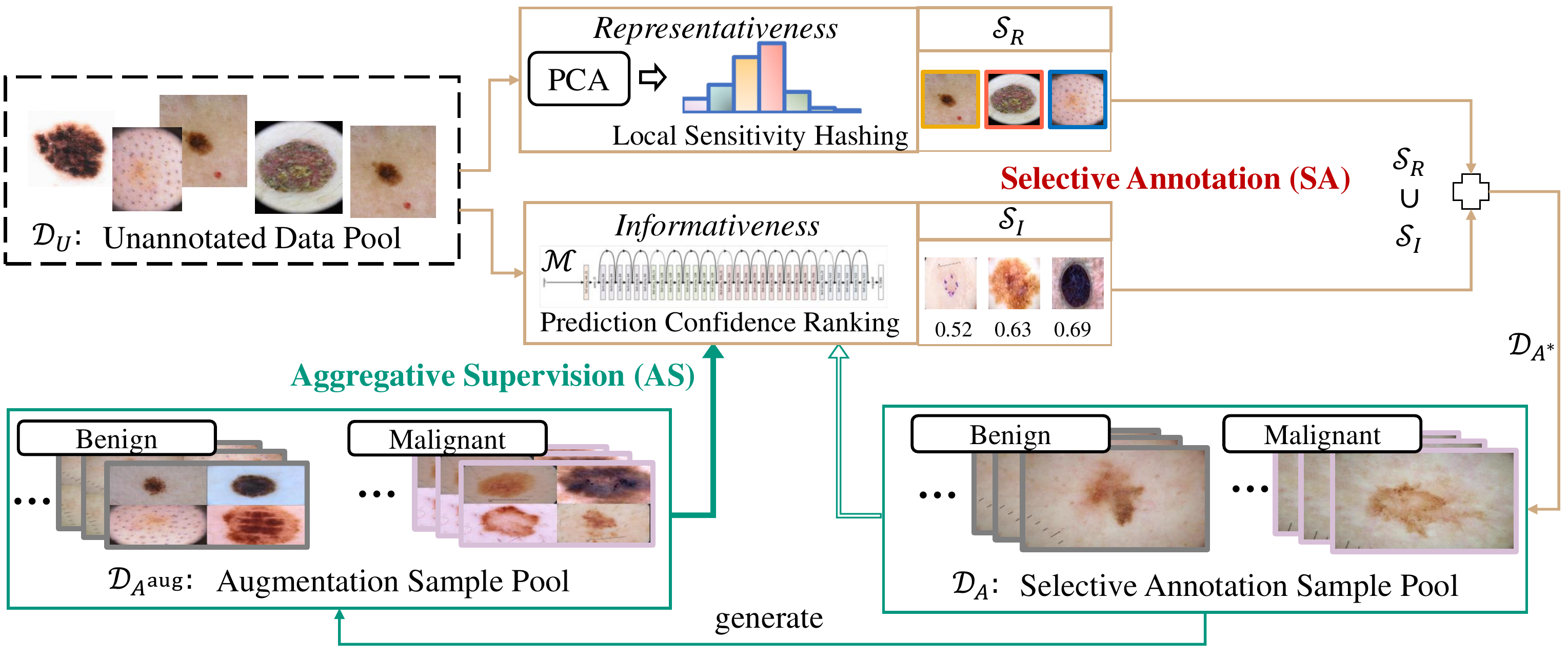}
    \caption{Overview of our proposed active learning framework for skin lesion analysis. In each iteration, from unannotated data pool $\mathcal{D}_U$, we select a worthy-annotation set $\mathcal{D}_{A^{\ast}}$ composing representative samples $\mathcal{S}_R$ and informative samples $\mathcal{S}_I$. Moreover, we generate augmentations $\mathcal{D}_{A^{\text{aug}}}$ of all the gathered annotated data pool $\mathcal{D}_A$.
    Finally, the model is updated by supervised learning with $\mathcal{D}_A \cup \mathcal{D}_{A^{\text{aug}}}$.}
    \label{fig:fram}
\end{figure}

\section{Methods}
Our framework is illustrated in Fig.~\ref{fig:fram}.
We first train the ResNet-101 model $\mathcal{M}$ with the annotated set of $\mathcal{D}_A \!=\! \{(x_j,y_j)\}_{j=1}^{Q}$, which is initialized with randomly selected 10\% data from the unlabelled sample pool $\mathcal{D}_U\!=\! \{x_i\}_{i=1}^{T}$. Next, we iteratively selecting samples, aggregating samples, and updating the model. 

\subsection{Selective Annotation (SA) with Dual-Criteria}
We select samples considering both criteria of informativeness and representativeness.
%\textcolor{red}{, with the ambition of enhancing model performance and covering sample distribution}. 
The informativeness is calculated based on the prediction of the trained model. The representativeness is obtained by PCA features and hashing method.
In our framework, we call this procedure as \textit{selective annotation (SA)}.

Firstly, we test the unlabelled samples with the current trained model. The images with low prediction confidences computed from the model are selected as informative ones, since they are nearby the decision boundary.
The model would present relatively lower confidence when encountering those new ``hard'' unlabelled samples,
which usually have either ambiguous pattern or rare appearance.
For each sample, the highest prediction probability across all classes, is regarded as its model certainty.
With ranking $\mathcal{D}_U$ according to certainties, the lower certainty indicates stronger informativeness. The selected samples following this aspect of criteria are represented as $\mathcal{S}_I$:
\begin{equation}
\mathcal{S}_I \gets \underset{x_i}{\operatorname{Rank}}(\{ \mathcal{M}(x_i) \}, N_I),
\label{eqn:SI}
\end{equation}
where $\mathcal{M}(x_i)$ is certainty of current model $\mathcal{M}$ for each sample $x_i$ in $\mathcal{D}_U$, ranking is in ascending order, and the first $N_I$ samples are selected.
We set $N_I = 10\% \!\times\! N \!\times\!  \gamma$ where $N$ is the total number of available samples, and $\gamma$ is the sample selection ratio of informativeness criterion. 10\% is the hyper-parameter which controls the scale of newly selected samples during each round of sample selection.

Next, considering sample diversity, we desire the added samples present dissimilar appearances, and hence, are representative for the entire dataset.
Specifically, we regard feature-level difference as an indicator of sample diversity.
To avoid using features from the same CNN as used for informative sample selection, we compute the first principal component of the image as features for data diversity.
With the PCA features, we map similar unlabelled items into the same buckets using local sensitivity hashing (LSH), which is for efficient approximate nearest neighbor search.
Next, we uniformly fetch samples from each bucket and obtain the set of $\mathcal{S}_R$ as representative samples. This process is formulated as:
\begin{equation}
\mathcal{S}_R \gets \underset{x_i}{\operatorname{UniSample}}(LSH(\{PCA(x_i)\},K),N_R),
\label{eqn:SR}
\end{equation}
where $K\!=\!10$ is our number of buckets in LSH. We set $N_R=10\% \!\times\! N \!\times\!  (1-\gamma)$ with $(1-\gamma)$ being the sample selection ratio of representativeness criterion.
As the PCA features are independent of the learned CNN, our obtained $\mathcal{S}_R$ and $\mathcal{S}_I$ are decoupled and highly complementary. With one round of SA, we get the additional labelled set of samples as $\mathcal{D}_{A^{\ast}} \! \! = \!  \mathcal{S}_I \cup \mathcal{S}_R $ and update $\mathcal{D}_A \gets \mathcal{D}_A \!\cup \mathcal{D}_{A^\ast}$.

\subsection{Aggregative Supervision (AS) with Intra-class Augmentation}
In active learning, majority previous efforts have focused on how to select samples, but somehow neglected how to effectively harness them to produce more distinguishable features.
As usually the selected training samples are very challenging and ambiguous, 
it is important to design strategies which can sufficiently unleash the potential values of these newly labelled samples.
If just directly using such samples to fine-tune the model, we may encounter high risks of over-fitting, since the updated decision boundary would be curly to fit the ambiguous images.
To enhance the model's capability to deal with those ambiguous samples, we propose to aggregate the images into new form of augmented samples to update the model. In our framework, we call this procedure as \textit{aggregative supervision (AS)}.

\begin{figure}[t!]
	\centering
	\includegraphics[width=0.7\linewidth]{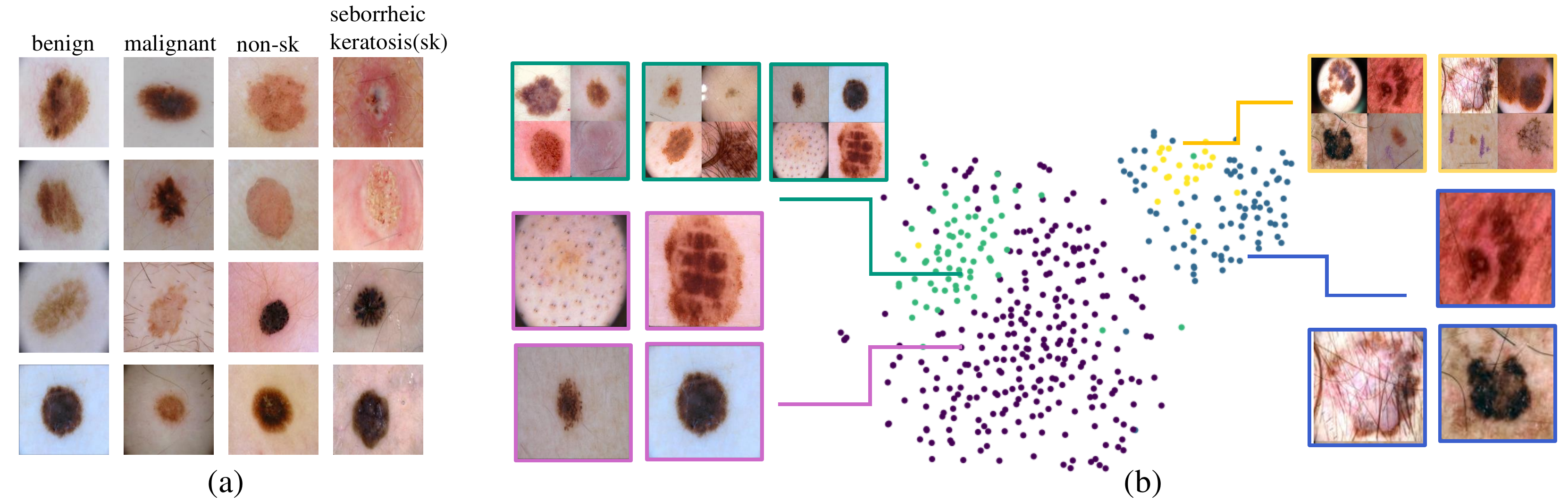}
	\caption{(a) Skin lesion images with limited inter-class variance while large intra-class variance. (b) Embedding of high-level CNN features of the original and augmented samples using t-SNE. The purple and green dots are original and augmented benign data. The blue and yellow dots are original and augmented malignant data.
	Augmented samples are natural clustering with original ones in the high-level semantic space.}
	%\textcolor{red}{We test these non-trained data on a trained model, for presenting the natural clustering between aggregated samples and original samples in the high-level semantic space, and demonstrating the feature-level consistency between them.}}
	\label{fig:fitting}
\end{figure}

Specifically, we aggregate images from the same class in pixel space, by stitching four intra-class images in a $2 \! \times \! 2$ pattern, as presented in Fig.~\ref{fig:fitting}~(b).
Such a concatenation of samples from the same class can provide richer yet more subtle clues for the model to learn more robust features to reduce intra-class variance, especially given the highly ambiguous and limited number of samples obtained from SA process.
In a sense that the model aims to discriminate between distributions of benign and malignant images,
the proposed sample aggregation scheme can be beneficial to reduce the influence of individual complicated sample on the model, and percolate the underlying pattern inherent in each category.
Finally, the aggregated image is resized to the same size as the original resolution, and its label is the same class of those composed images.
Generally, our strategy shares the pixel-level augmentation spirit as mix-up~\cite{zhang2017mixup}, while we can avoid overlapping the ambiguous contents of inter-class images with limited appearance difference.

To demonstrate the effectiveness of the proposed aggregation scheme at feature level,
we embed the CNN features of the original images and the aggregated images with our intra-class stitching
onto a 2D plane using t-SNE, see Fig.~\ref{fig:fitting}.
We employ the features obtained from the last fully connected layer (before softmax), as these features have strong semantic meanings.
Note that these aggregated samples haven't yet been used to train the model.
We observe that the aggregated samples naturally group together with the ordinary images within the class, when mapped into the higher-level space with a pre-learned feature extractor (i.e., the CNN model).
This demonstrates that our aggregation scheme can provide a new and informative form of training images, offering apparently different view in raw pixel space while maintaining the essential patterns of its category in the highly-abstracted semantic space.

\section{Experimental Results}
\textbf{Dataset.}
We extensively validate our proposed active learning framework on two different tasks using the public data of \textit{ISIC 2017 Skin Lesion Classification Challenge}~\cite{codella2018skin}.
These two tasks hold different aspects of challenges and sample ambiguity characteristics.

Same as the state-of-the-art methods on the leaderboard~\cite{diaz2018dermaknet,EResNet}, in addition to the 2,000 training images provided by the challenge, we acquired 1,582 more images (histology or expert confirmed studies) from the ISIC archive~\cite{codella2018skin} to build up our available database.
In total, we got 3,582 labelled images (2,733 benign and 849 malignant) as our training data pool. We directly utilized the validation set (150 images) and test set (600 images) of the ISIC challenge.
\\
\\
\textbf{Implementations.}
The luminance and color balance of input images are normalized exploiting color constancy by gray world. Images are resized to 224$ \times $224 to match input size of pre-trained ResNet-101 model. The images are augmented with rotating by up to 90$^\circ$, shearing by up to 20$^\circ$, scaling within [0.8, 1.2], and random flipping horizontally and/or vertically.
We use weighted cross-entropy loss with Adam optimizer and initial learning rate as 1e-4. Code will be released.
\\
\\
\textbf{Evaluation metrics.} For quantitative comparisons, our evaluations followed the challenge given metrics, which consist of accuracy (ACC), area under ROC curve (AUC), average precision (AP), sensitivity (SE) and specificity (SP).
\begin{table}[t]
	\caption{Quantitative evaluations of our proposed active learning framework for skin lesion analysis on two different classification tasks.}
	\resizebox{1\textwidth}{!}
	{
		\begin{tabular}{|l|l|c|c|c|c|c|c|c|c|c|c|c|c|}
\hline
\multicolumn{2}{|c|}{}                          &                                                                            &                                                                          & \multicolumn{5}{c|}{Task1}                                                                                                                                                                                                                                  & \multicolumn{5}{c|}{Task2}                                                                                                                                                     \\ \cline{5-14}
\multicolumn{2}{|c|}{\multirow{-2}{*}{Methods}} & \multirow{-2}{*}{\begin{tabular}[c]{@{}c@{}}Data \\ Amount \end{tabular}} & \multirow{-2}{*}{\begin{tabular}[c]{@{}c@{}}Extra \\ Label\end{tabular}} & ACC                                   & AUC                                   & AP                                                         & SE                                                & SP                                                         & ACC           & AUC                                   & AP                                    & SE                                    & SP                                    \\ \hline
                               & Monty~\cite{diaz2018dermaknet}          & 100\%                                                                      & ${\surd}$                                                                      & 0.823                                 & {\color[HTML]{333333} 0.856}          & \multicolumn{1}{l|}{{\color[HTML]{333333} 0.654}}          & \multicolumn{1}{l|}{{\color[HTML]{333333} 0.103}} & \multicolumn{1}{l|}{{\color[HTML]{333333} \textbf{0.998}}} & 0.875          & {\color[HTML]{333333} \textbf{0.965}} & {\color[HTML]{333333} \textbf{0.839}} & {\color[HTML]{333333} 0.178}          & \textbf{0.998}                        \\
                               & Popleyi~\cite{EResNet}        & 100\%                                                                      & $\times$                                                                       & 0.858                                 & \textbf{0.870}                        & \textbf{0.694}                                             & 0.427                                             & 0.963                                                      & \textbf{0.918} & 0.921                                 & 0.770                                 & 0.589                        & 0.976                                 \\
\multirow{-3}{*}{Leaderboard} & Full-data (ResNet-101)     & 100\%                                                                      & $\times$                                                                       & {\color[HTML]{333333} \textbf{0.863}} & {\color[HTML]{333333} 0.821}          & {\color[HTML]{333333} 0.590}                               & {\color[HTML]{333333} \textbf{0.496}}             & {\color[HTML]{333333} 0.952}                               & 0.903          & {\color[HTML]{333333} 0.941}          & {\color[HTML]{333333} 0.773}          & {\color[HTML]{333333} \textbf{0.856}} & 0.912                                 \\ \hline
                               & Random (Rand)           & 50\%$/$40\%                                                                       & $\times$                                                                       & 0.825                                 & 0.795                                 & {\color[HTML]{333333} 0.520}                               & 0.359                                             & 0.934                                                      & 0.878          & {\color[HTML]{333333} 0.923}          & 0.731                                 & 0.722                                 & 0.906                                 \\
                               & AIFT~\cite{zhou2017fine}           & 50\%$/$40\%                                                                       & $\times$                                                                       & 0.810                                 & 0.754                                 & 0.447                                                      & \textbf{0.385}                                    & 0.913                                                      & 0.885          & {\color[HTML]{000000} 0.907}          & 0.677                                 & 0.711                                 & {\color[HTML]{333333} \textbf{0.916}} \\
\multirow{-3}{*}{~~Selection}           & SA (Ours)      & 50\%$/$40\%                                                                       & $\times$                                                                       & {\color[HTML]{333333} \textbf{0.847}} & \textbf{0.800}                        & {\color[HTML]{000000} \textbf{0.575}}                      & 0.368                                             & {\color[HTML]{333333} \textbf{0.963}}                      & \textbf{0.903} & {\color[HTML]{333333} \textbf{0.938}} & {\color[HTML]{333333} \textbf{0.784}} & {\color[HTML]{333333} \textbf{0.844}} & {\color[HTML]{333333} 0.914}          \\ \hline
                               & SA (Ours)+Mix-up~\cite{zhang2017mixup} & 50\%$/$40\%                                                                       & $\times$                                                                       & 0.467                                 & 0.572                                 & {\color[HTML]{333333} 0.273}                               & {\color[HTML]{333333} \textbf{0.615}}             & 0.431                                                      & 0.720          & {\color[HTML]{333333} 0.638}          & 0.361                                 & 0.124                                     & 0.824                            \\
\multirow{-2}{*}{Aggregation}       & SA+AS (\textbf{Ours})    & 50\%$/$40\%                                                                    & $\times$                                                                       & {\color[HTML]{333333} \textbf{0.860}} & {\color[HTML]{333333} \textbf{0.831}} & \multicolumn{1}{l|}{{\color[HTML]{333333} \textbf{0.600}}} & \multicolumn{1}{l|}{0.479}                        & \multicolumn{1}{l|}{{\color[HTML]{333333} \textbf{0.952}}} & \textbf{0.908} & \textbf{0.934}                        & \textbf{0.755}                        & \textbf{0.756}                        & {\color[HTML]{333333} \textbf{0.935}}          \\ \hline
\end{tabular}
	}
	\label{table:2taskcompare}
\end{table}

\subsection{Results of Cost-effective Skin Lesion Analysis}
In our active learning process, based on the initially randomly selected 10\% data, we iteratively added training samples until obtaining predictions which cannot be significantly improved ($p \! > \!0.05$) over the accuracy of last round. It turns out that we only need 50\% of the data for Task-1 and 40\% of the data for Task-2. 

The overall performance for Task-1 and Task-2 are representatively presented in Fig.~\ref{fig:finalcompare}\textcolor{blue}{(a)} and Fig.~\ref{fig:compare_task2}\textcolor{blue}{(a)}.
In Fig.~\ref{fig:finalcompare}\textcolor{blue}{(a)}, we present the baseline of active learning which is random sample selection (purple). By using our proposed dual-criteria sample selection (green), the accuracy gradually increases and keeps higher than the baseline through different query ratios. Further using our aggregative supervision (red), the accuracy achieves $86.0\%$ when using only $50\%$ samples, which is very close to the accuracy of $86.3\%$ with full-data training (yellow).
In Fig.~\ref{fig:compare_task2}\textcolor{blue}{(a)}, by actively querying worth-labelling images, our proposed method can finally exceed the performance of full-data training only using 40\% samples.
In addition, when comparing with the state-of-the-art method of AIFT (blue)~\cite{zhou2017fine}, our proposed method can outperform it consistently across all sample query ratios on both tasks. This validates that our deriving dual-criteria in a decoupled way is better than only relying on currently learned network.

In Table~\ref{table:2taskcompare},  we categorize the different comparison methods into three groups, 
%\textcolor{red}{(1)  We compare our full data trained model (Resnet-101)  with the leading methods in the challenge. (2) We compare the proposed sample selection strategy (SA) among existing active learning methods. (3) Based on our SA strategy, we further compare the stiching augmentation strategy (AS) with mix-up.}
i.e., the leading methods in challenge, active learning only with sample selection strategy, and further adding the sample augmentation strategy.
The amount of employed annotated data is indicated in \emph{data amount} column.
For leaderboard, only rank-2~\cite{diaz2018dermaknet} and rank-4~\cite{EResNet} methods are included, as rank-1 method used non-image information (e.g., sex and age) and rank-3 method used much more extra data besides the ISIC archive ones.
Nevertheless, we present the challenge results for demonstrating the state-of-the-art performance of this dataset.
We focus on active learning part, with our implemented full-data training as standard bound.
From the Table~\ref{table:2taskcompare}, we see that our SA can outperform AIFT~\cite{zhou2017fine}, and AS can outperform mix-up~\cite{zhang2017mixup}, across almost all evaluation metrics on both tasks. Overall, our proposed method achieves highly competitive results against full-data training and challenge leaderboard, with significantly cost-effective labellings.
\subsection{Analysis of Components in Active Learning Framework}
\begin{figure}[t!]
    \centering
    \includegraphics[width=0.95\linewidth]{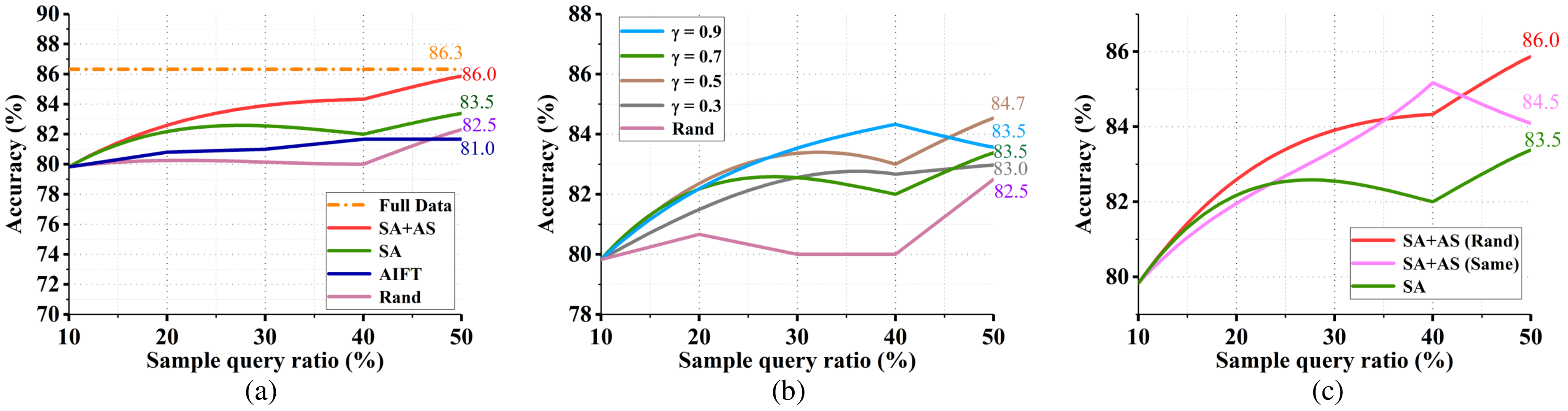}
    \caption{Experimental results of our proposed active learning framework on Task 1. (a) Overall accuracy of different methods at sample query ratios. (b) Ablation study of SA, by adjusting $\gamma$. (c) Ablation study of AS, by changing the choice of stitched images.}
    \label{fig:finalcompare}
\end{figure}

\begin{figure}[t!]
	\centering
	\includegraphics[width=0.95\linewidth]{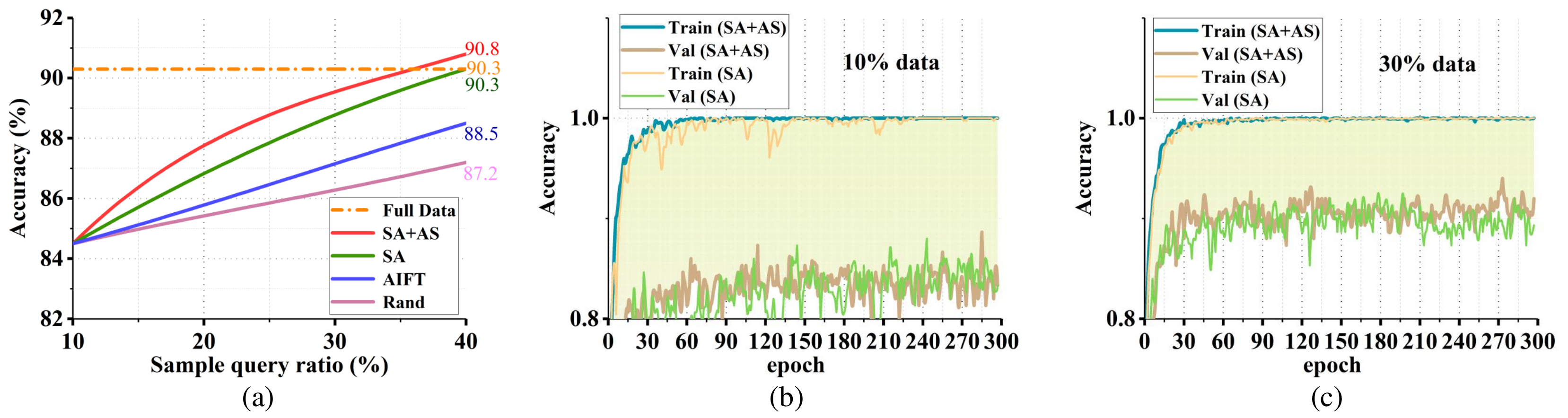}
	\caption{Experimental results of our proposed active learning framework on Task 2. (a) Overall accuracy of different methods at sample query ratios. (b)-(c) Observation of narrowing generalization gap and alleviating over-fitting by our active learning method.}
	\label{fig:compare_task2}
\end{figure}

Firstly, we investigate the impact of hyper-parameter setting in sample selection.
We adjust the ratio between $\mathcal{S}_I$ and $\mathcal{S}_R$ by changing the $\gamma$, as shown in Fig.~\ref{fig:finalcompare}\textcolor{blue}{(b)}.
Varying the ratio $\gamma$ would bring fluctuation on performance, but even the worst case is much better than random selection.
We choose to use $\gamma=0.7$ as the basis AS process, since it reflects the average-level performance of the SA step.
%\textcolor{red}{More interestingly, these lines all have some degree of performance decrease except $\gamma=0.3$ (representative surpass informative), which verifies the complementary effectiveness of representative selection.}

Secondly, we investigate the practically effective manner to stitch the intra-class samples.
As shown in Fig.~\ref{fig:finalcompare}\textcolor{blue}{(c)}, we compare stitching four randomly selected intra-class images and replicating the same image by four times.
For aggregative supervision, stitching different intra-class images can outperform replicating the same image, which exactly reflects that our designed augmentation strategy can help to improve performance by suppressing intra-class variance and sample ambiguity.

Finally, the results in Fig.~\ref{fig:compare_task2}\textcolor{blue}{(b)} and Fig.~\ref{fig:compare_task2}\textcolor{blue}{(c)} show that our proposed augmentation strategy can alleviate overfitting during model training.
The shadow area indicates the generalization gap between the training and validation sets. It unsurprisingly decreases with increasing the data amount from $10\%$ to $30\%$. With more careful observation, we find that the SA+AS can generally surpass pure SA on validation set, which demonstrates the effectiveness of alleviating over-fitting using our augmented new-style samples.

\section{Conclusion}
This paper presents a novel active learning method for annotation cost-effective skin lesion analysis.
We propose a dual-criteria to select samples, and an intra-class sample aggregation scheme to enhance the model.
Experimental results demonstrate that using only up to 50\% of the labelled samples, we can achieve the state-of-the-art performance on two different skin lesion analysis tasks.
\begin{comment}
The proposed methods are general enough to be harnessed in other similar tasks, where how to reduce the annotation workload while maintaining model performance plays a key role in success.
\end{comment}
\\
\\
\textbf{Acknowledgments.}
The work described in this paper was supported by the 973 Program with Project No.2015CB351706, the National Natural Science Foundation of China with Project No.U1613219 and the Hong Kong Innovation and Technology Commission through the ITF ITSP Tier 2 Platform Scheme under Project ITS/426/17FP.

\bibliographystyle{splncs04}
\bibliography{ref}

\begin{thebibliography}{10}
\providecommand{\url}[1]{\texttt{#1}}
\providecommand{\urlprefix}{URL }
\providecommand{\doi}[1]{https://doi.org/#1}

\bibitem{codella2018skin}
Codella~{et al.}, N.C.: Skin lesion analysis toward melanoma detection: A
  challenge at the 2017 international symposium on biomedical imaging (isbi),
  hosted by the international skin imaging collaboration (isic). In: ISBI. pp.
  168--172 (2018)

\bibitem{diaz2018dermaknet}
Diaz, I.G.: Dermaknet: Incorporating the knowledge of dermatologists to
  convolutional neural networks for skin lesion diagnosis. IEEE Journal of
  Biomedical and Health Informatics pp. 547--559 (2018)

\bibitem{esteva2017dermatologist}
Esteva, Andre~{et al.}, K.B.: Dermatologist-level classification of skin cancer
  with deep neural networks. Nature  \textbf{542}(7639), ~115 (2017)

\bibitem{huang2010active}
Huang, S.J., Jin, R., Zhou, Z.H.: Active learning by querying informative and
  representative examples. In: NIPS. pp. 892--900 (2010)

\bibitem{EResNet}
Lei, B., Jinman, K., Euijoon, A., Dagan, F.: Automatic skin lesion analysis
  using large-scale dermoscopy image and deep residual networks.
  arXiv:1703.04197  (2017)

\bibitem{mahapatra2018efficient}
Mahapatra, D., Bozorgtabar, B., Thiran, J.P., Reyes, M.: Efficient active
  learning for image classification and segmentation using a sample selection
  and conditional generative adversarial network. In: MICCAI. pp. 580--588
  (2018)

\bibitem{settles.tr09}
Settles, B.: Active learning literature survey. Computer Sciences Technical
  Report~1648, University of Wisconsin--Madison (2009)

\bibitem{xue2019robust}
Xue, C., Dou, Q., Shi, X., Chen, H., Heng, P.A.: Robust learning at noisy
  labeled medical images: Applied to skin lesion classification. ISBI  (2019)

\bibitem{yang2017suggestive}
Yang, L., Zhang, Y., Chen, J., Zhang, S., Chen, D.Z.: Suggestive annotation: A
  deep active learning framework for biomedical image segmentation. In: MICCAI.
  pp. 399--407 (2017)

\bibitem{zhang2017mixup}
Zhang, H., Cisse, M., Dauphin, Y.N., Lopez-Paz, D.: mixup: Beyond empirical
  risk minimization. ICLR  (2017)

\bibitem{zhang2018skin}
Zhang, J., Xie, Y., Wu, Q., Xia, Y.: Skin lesion classification in dermoscopy
  images using synergic deep learning. In: MICCAI. pp. 12--20 (2018)

\bibitem{aaai2019ra}
Zheng, H., Yang, L., Chen, J., Han, J., Zhang, Y., Liang, P., Chen, D.Z.,
  et~al.: Biomedical image segmentation via representative annotation. In: AAAI
  (2019)

\bibitem{zhou2017fine}
Zhou, Z., Shin, J.Y., Zhang, L., Gurudu, S.R., Gotway, M.B., Liang, J.:
  Fine-tuning convolutional neural networks for biomedical image analysis:
  Actively and incrementally. In: CVPR. pp. 4761--4772 (2017)

\end{thebibliography}
\end{document}